# Artificial Neural Network for Performance Modeling and Optimization of CMOS Analog Circuits

Mriganka Chakraborty
Assistant Professor
Department Of Computer Science & Engineering, Seacom Engineering College
West Bengal, India.

## ABSTRACT
This paper presents an implementation of multilayer feed forward neural networks (NN) to optimize CMOS analog circuits. For modeling and design recently neural network computational modules have got acceptance as an unorthodox and useful tool. To achieve high performance of active or passive circuit component neural network can be trained accordingly. A well trained neural network can produce more accurate outcome depending on its learning capability. Neural network model can replace empirical modeling solutions limited by range and accuracy.[2] Neural network models are easy to obtain for new circuits or devices which can replace analytical methods. Numerical modeling methods can also be replaced by neural network model due to their computationally expansive behavior.[2][10][20]. The proposed implementation is aimed at reducing resource requirement, without much compromise on the speed. The NN ensures proper functioning by assigning the appropriate inputs, weights, biases, and excitation function of the layer that is currently being computed. The concept used is shown to be very effective in reducing resource requirements and enhancing speed.

## Keywords
Artificial Neural Network, CMOS, Analog Circuit Optimization.

## 1. INTRODUCTION
Neural networks, are also known as artificial neutral network (ANN's), are information processing system with their design inspired by the studies of the ability of the human brain to learn from observations and to generalize by abstraction.[2] The fact that neural network can be trained to learn any arbitrary nonlinear input/output relationships from corresponding data and the acquired knowledge has resulted in their use in a number of areas such as pattern recognition, speech processing,[2][4] control, bio medical engineering, RF and microwave etc. recently, (ANNs) have been applied to CMOS analog circuit design and optimization problems as well. Neural networks are first trained to model the electrical behavior of passive and active components/ circuits.[2][7] These trained neural networks, often referred to as neural-network models, can then be used in high level simulation and design, providing first and accurate answers to the task they have learnt by acquiring the knowledge from their training. Neural networks are effective and efficient alternatives to conventional method such as numerical modeling methods, which could be highly computationally expensive, or analytical methods which could be difficult to obtain for newly achived devices, or empirical modeling solutions due to huge range and limited accuracy.[2][6] Neural network techniques have been used for a very wide variety of applications and modeling methods.[5] An analog system is typically characterized by a set of performance parameters used to succinctly quantify the properties of the circuit given fixed topology; circuit synthesis is the process of determining numerical values for all components in the circuit such that the circuit conforms to a set of performance constraints. Due to the high degree of nonlinearity and interdependence among design variables, manual design of an analog circuit is often reduced to a process of trial and error in which the solution space is searched in an ad hoc manner for a circuit satisfying all constraints.[7] The numerical circuit simulator SPICE is often used as a bench mark of comparison to determine the relative accuracy of alternative schemes for evaluating the performance of analog circuits. However the computational requirements of running SPICE limit its use when attempting to evaluate a circuit's performance parameters during circuit synthesis. Stochastic combinatorial optimization methods require the computation of performance parameters for a large number of circuit sizing alternatives[6]. It is therefore beneficial to reduce the time associated with generating performance estimates. Neural network models are used to provide robust and accurate estimates of performance parameters for several CMOS analog circuits[18][19][20]. A neural network of sufficient size can estimate functional mappings to an arbitrary precision given a finite discrete set of training data. Hyper dimensional non-linear functions are readily modeled using neural networks. Neural networks can also easily incorporate knowledge of system behavior. Course functional models can be embedded in the network structure reducing the functional complexity that must be mapped by the network. These often results in a smaller network size and reduction in training effort.[9] Once trained with a particular functional mapping, the evaluation time of a neural model is very fast. However training algorithms can help reduce the interaction needed to determine and appropriate network size. The evaluation time for the neural models is much less than that required by a full SPICE simulation, the models can be incorporated into a circuit synthesis algorithm used to optimize a fitness function based on performance parameter constraints. Neural networks have recently gained attention as a fast, accurate and flexible tool to modeling, simulation and design. Each time a new network is trained, or an old network is retrained, the shape of the function described by the neural model changes, complicating the issue of where to place additional sample points.[15] The neural network models provide a great deal of time savings in situations where a fixed topology must be reused and re-synthesized many times which is the primary target for modeling and synthesis of analog circuits using neural network models.[10]

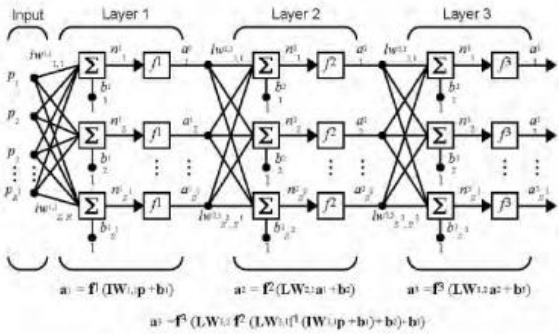

**Fig 1: Multi layer perception neural network with input layer and multiple hidden layers.**

## 2. METHEDOLOGY
## 2.1 Problem formulation and input data processing
### 2.1.1 Inputs and outputs of Neural Network
The first step for developing a neural network model is the identification of proper inputs and outputs parameters. The output parameters are determined based on the purpose and the functionality of the neural network model. Other factors influencing the type of outputs parameters are 1) ease of data generation, 2) ease of incorporation of the neural model into circuit simulators, etc.. Neural model input parameters are those device/circuits parameters that affect the output parameter values accordingly. Input parameters can be selected depending on various kind of function which can involve neural network model for achieving more accurate output at a higher speed.

### 2.1.2 Range of input data with sample distribution
The next thing is to define the specific range of data to be used in ANN model development and the distribution of samples within that specified range. Training data is sampled slightly beyond the model utilization range as a measure to ensure reliability of the neural model at the boundaries of model utilization range. Once the range of input parameters is defined, a sampling distribution needs to be chosen. Commonly used sample distributions include uniform grid distribution, non-uniform grid distribution, design of experiment methodology, star distribution and random distribution process. In uniform and non-uniform grid distribution, each input parameter is sampled at equal and unequal intervals accordingly. This is useful when the problem behavior is highly non-linear in certain regions and sub regions of the space and dense sampling is needed in such regions and sub regions. Sample distribution base on design of experiment and star distribution are used in situations where training data generation is highly expensive.

### 2.1.3 Generating input data
Sample pairs are then generated using either by simulation software or by measurement setup. The generated dataset could be used for training the neural network and testing the resulting neural network model. In practice, both simulations and measurements setup could generate small errors. While errors in simulation could be due to truncation or round off or non-convergence, errors in measurement could be due to equipment limitations or tolerance limits. Data generation is defined as the use of simulation or measurement to obtain sample pairs.[2][16]  The total number of samples is chosen such that the developed neural network model best represents the given problem. A general guideline is to generate larger number of samples for a nonlinear high-dimensional problem and fewer samples for a relatively smooth low dimensional problem due to accuracy measure.

### 2.1.4 Input data organization and processing
The generated sample pairs could be divided into three sets, namely, training data set, validation data set and test data set. Training data is used to guide the training process i.e. to update the neural network weight parameters during training and to acquire required knowledge for the provided problem. Validation data is used to monitor the quality of the neural network model during training and to determine stop criteria for the training process of the neural network model.[2][22] Test data is used to independently examine the final quality of the trained neural network model in terms of accuracy and generalization capability and processing speed. Contrary to binary data in pattern recognition applications, the orders of magnitude of various inputs and outputs parameter values can be very much different from one another.[2] As such, a systematic preprocessing and biasness removal of training data called scaling is mostly desirable for efficient neural network training program.

## 2.2 Training of Neural Network
### 2.2.1 Initialization of weight parameters
Preparing the neural network for training process. The neural network weight parameters are initialized so as to provide a good starting point for training process. The widely used strategy for MLP weight initialization is to initialize the weights with small range of random values. Another method suggest that the range of random weights be in aversely proportional to the square root of number of stimuli a neuron receives on average rate.[2] To improve the convergence of training method, one can use a variety of distributions and/or different ranges and different variance for the random number generators used in initialization process of the ANN weights.

### 2.2.2 Creation of training process
The most important step in neural model development is the neural network training process as it provides the knowledge for future performance of the neural network model for specific applications. The training data consists of sample pairs and some vectors representing the inputs and expected outputs of the neural network model. The neural network training process can be categorized into batch mode training and sample by sample training. In sample by sample training, also called online training. In batch mode training, also known as offline training, weight and epoch is define as a stage of the training process that involves presentation of all the training data to the neural network once.[2][9][21]

### 2.2.3 Error computation
The error of the neural network can be determined after training and simulation with the provided data set. The difference between actual data and simulated test data is divided by the total number of dataset to get the error margin of the network. Mean square error can also be calculated for the neural network model. The desired type of error for different type of function can be easily achived from the neural network model.

## 2.3 Implementation of Neural Network Model

First it is needed to select a development environment where the neural model should be developed and implemented. Therefore MATLAB is being selected to design and develop the required neural network model. The neural network is developed by generating proper program code it the MATLAB console and then decorated with proper algorithms and required problem definitions. Multi layer perception Neural Network model has been created for more accuracy with more numbers of hidden layers for quality measure.

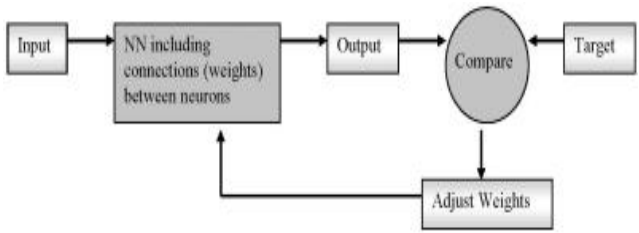

**Fig 2: Pictorial view of neural network training and testing.**

### 2.3.1 Working with general function

The next step is to supplying proper data to the neural network in order to a particular problem set. First the problem definition is inserted in the program code for formulation of learning process. Here a simple a general equation is used as the problem set i.e. [out=a.*(x).^2+b.*(x).*(y)+c.*(y).^2]. This equation can be described as follows :

$$S = ax^2 + 2bxy + cy^2$$

Where S is the calculated sum of the equation and a, b, c are the constant terms. Here the neural network input layer is having two inputs x and y which is a collection of data set randomly created by the program code. But it is necessary to map the proper index of those input data sets with one another after properly shuffling the data set in a coherent orderly manner to remove the complete biasness of the data set. Here input sets x and y are randomly shuffled maintaining proper index mapping and to remove the data redundancy [21].

The next approach is to select the training and testing data sets from the given data sets. The 80 percent of the total data set is taken as the training dataset and rest of it as the test data set by trimming the whole data set into two different portion maintaining the proper index mapping. With the training data set the neural network is trained and with the test data set the performance evaluation of the neural network is being processed.

The next step was to train the developed neural network with the provided training data set. The minimum and the maximum value is processed from the provided training dataset for defining the minimum and maximum value of the neural network while training for better performance. Next the training process is processed by providing proper training function and parameters for the particular problem set and the parameters of the neural network is also processed with some particular parameters i.e. proper algorithm, number of neurons, number of layers, number of hidden layers etc.. Then the neural network is initialized with those provided parameters for iterative estimation and better performance. It is possible to check the out comes of the neural network training for reducing the error factor of the network. Choosing proper algorithm for training optimizes the time of neural network training which is a crucial factor. Feed Forward Back Propagation Network is created with Levenberg-Marquardt algorithm to minimize the training latency. The calculated CPU time is achieved to identify the performance of the neural network.

Hence the neural network learned to compute the given problem with some particular data set and acquired the knowledge of solving that particular problem using other data set provided. Next simulation of the neural network with test data set is processed.

Lastly the calculated error of the neural network is achieved after neural network simulation. The error E is calculated as:

$$E = \{1/n \sum_{i=1}^{n}[\frac{|atdv - stdv|}{atdv}]\} \times 100$$

where, atdv is actual test data value and stdv is simulated test data value, n is the total number of data and i is the number of iterations. More and more training with training data sets will improve the accuracy level of the neural network and will optimize the error.

### 2.3.2 Working with other standard functions

We have also simulated the neural network with some of the standard unconstrained test problem functions i.e. Beale function, Booth function, Bohachevsky function, Easom function, Hump function.

### 2.3.2.1 Working with Bohachevsky function-1

First the problem definition is inserted in the program code for formulation of learning process i.e.

$B_1(X_1, X_2) = X_1^2 + 2X_2^2 - 0.3\cos(3\pi X_1) - 0.4\cos(4\pi X_2) + 0.7$     Where $B_1$ is the calculated sum of the equation. Here the neural network input layer is having two inputs $X_1$ and $X_2$ which is a collection of data set randomly created by the program code within a given domain i.e. $-50 \leq X_i \leq 100$, where i=1,2…. But it is necessary to map the proper index of those input data sets with one another after properly shuffling the data set in a coherent orderly manner to remove the complete biasness of the data set. Here input sets $X_1$ and $X_2$ are randomly shuffled maintaining proper index mapping and to remove the data redundancy.

*2.3.2.2 Working with Bohachevsky function-2*

First the problem definition is inserted in the program code for formulation of learning process i.e.

$$B_2(X_1, X_2) = X_1^2 + 2X_2^2 - 0.3\cos(3\pi X_1)\, 0.4\cos(4\pi X_2) + 0.3$$

Where $B_2$ is the calculated sum of the equation. Here the neural network input layer is having two inputs $X_1$ and $X_2$ which is a collection of data set randomly created by the program code within a given domain i.e. $-50 \leq X_i \leq 100$, where i=1,2…. But it is necessary to map the proper index of those input data sets with one another after properly shuffling the data set in a coherent orderly manner to remove the complete biasness of the data set. Here input sets $X_1$ and $X_2$ are randomly shuffled maintaining proper index mapping and to remove the data redundancy.

*2.3.2.3 Working with Beale function*

First the problem definition is inserted in the program code for formulation of learning process i.e.

$$B_L(X) = (1.5 - X_1 + X_1 X_2)^2 + (2.25 - X_1 + X_1 X_2^2)^2 + (2.625 - X_1 + X_1 X_2^3)^2$$

Where $B_L$ is the calculated sum of the equation. Here the neural network input layer is having two inputs $X_1$ and $X_2$ which is a collection of data set randomly created by the program code within a given domain i.e. $-4.5 \leq X_i \leq 4.5$, where i=1,2…. But it is necessary to map the proper index of those input data sets with one another after properly shuffling the data set in a coherent orderly manner to remove the complete biasness of the data set. Here input sets $X_1$ and $X_2$ are randomly shuffled maintaining proper index mapping and to remove the data redundancy.

*2.3.2.4 Working with Booth function*

First the problem definition is inserted in the program code for formulation of learning process i.e

$$B_O(X) = (X_1 + 2X_2 - 7)^2 + (2X_1 + X_2 - 5)^2$$

Where $B_O$ is the calculated sum of the equation. Here the neural network input layer is having two inputs $X_1$ and $X_2$ which is a collection of data set randomly created by the program code within a given domain i.e. $-4.5 \leq X_i \leq 4.5$, where i=1,2…. But it is necessary to map the proper index of those input data sets with one another after properly shuffling the data set in a coherent orderly manner to remove the complete biasness of the data set. Here input sets $X_1$ and $X_2$ are randomly shuffled maintaining proper index mapping and to remove the data redundancy.

*2.3.2.5 Working with Easom function*

First the problem definition is inserted in the program code for formulation of learning process i.e.

$$E_S(X_1, X_2) = X_1^2 + X_2^2 - \cos(18X_1) - \cos(18X_2)$$

Where $E_S$ is the calculated sum of the equation. Here the neural network input layer is having two inputs $X_1$ and $X_2$ which is a collection of data set randomly created by the program code within a given domain i.e. $-1 \leq X_i \leq 1$, where i=1,2…. But it is necessary to map the proper index of those input data sets with one another after properly shuffling the data set in a coherent orderly manner to remove the complete biasness of the data set. Here input sets $X_1$ and $X_2$ are randomly shuffled maintaining proper index mapping and to remove the data redundancy.

*2.3.2.6 Working with Hump function*

First the problem definition is inserted in the program code for formulation of learning process i.e.

$$H_M(X) = 4X_1^2 - 2.1X_1^4 + 0.33X_1^6 + X_1 X_2 - 4X_2^2 + 4X_2^4$$

Where $H_M$ is the calculated sum of the equation. Here the neural network input layer is having two inputs $X_1$ and $X_2$ which is a collection of data set randomly created by the program code within a given domain i.e. $-5 \leq X_i \leq 5$, where i=1,2…. But it is necessary to map the proper index of those input data sets with one another after properly shuffling the data set in a coherent orderly manner to remove the complete biasness of the data set. Here input sets $X_1$ and $X_2$ are randomly shuffled maintaining proper index mapping and to remove the data redundancy.

*2.3.3 Working with analog circuit function*

An analog circuit is developed in spice circuit editor having the equation

$$I_D = K`\omega/2L(V_{GS} - V_{TH})^2(1 + \lambda V_{DS})$$

Where $I_D$ Drain Current, $\omega$ is the effective channel width, L is the effective channel length, K' is the trans conductance parameter, $V_{GS}$ is the gate to source voltage, $V_{TH}$ is the threshold voltage, $V_{DS}$ is the drain to source voltage and $\lambda$ is the channel length modulation parameter of a MOS device. Here in linear region for different values of $V_{GS}$ and $V_{DS}$ there will be various values of $I_D$ .So for different values of $V_{GS}$ i.e. when $V_{GS}$ =0,1,2,3 there will be different values of $V_{GS}$ and $I_D$ which are retrieved from a spice synthesis report. Then the data chart is inserted into the code for training of neural network. After the training of the neural model it is then simulated with some data and it generated some value which is then compared with the actual spice data report and error is hence calculated.

First the problem definition is inserted in the program code for formulation of learning process.

$I_D = K`\omega/2L(V_{GS} - V_{TH})^2(1 + \lambda V_{DS})$

Where $I_D$ is the calculated sum of the equation. Here the neural network input layer is having two inputs $V_{GS}$ and $V_{TH}$ which is a collection of data set randomly selected from the spice output sheet. But it is necessary to map the proper index of those input data sets with one another after properly shuffling the data set in a coherent orderly manner to remove the complete biasness of the data set. Here input sets $V_{GS}$ and $V_{TH}$ are randomly shuffled maintaining proper index mapping and to remove the data redundancy.

## 3. EXPERIMENT RESULTS

**Table-1: Error and Time Calculation of ANN model For different functions.**

| Function Type | Neural Network Error | Time Taken (Seconds) |
|---|---|---|
| General Function (Lower number of neurons) | 0.59167% | 15.67 |
| General Function (Higher number of neurons) | 0.52124% | 1050.90 |
| Bohachevsky Function-1 | 4.6018% | 134.82 |
| Bohachevsky Function-2 | 0.67286% | 144.92 |
| Hump Function | 5.0895% | 177.56 |
| Beale Function | 2.2085% | 179.28 |
| Booth Function | 0.65626% | 171.67 |
| Analog Circuit Funcion-1 | 0.19039% | 9.56 |
| Analog Circuit Funcion-2 | 0.33277% | 4.06 |
| Analog Circuit Funcion-3 | 0.060012% | 3.76 |

The data generation process for different functions are considered for 500 sample set. Each sample set containing 400 training data set and 100 test date set. Analog circuit function is considered for 3 times with different configurations of voltage and current with different range basically extracted from spice circuit synthesis.

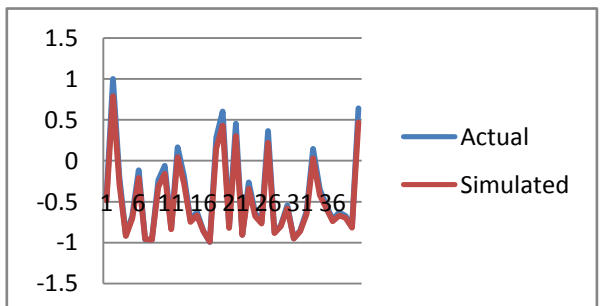

**Fig 3: Neural network output for Bohachevsky function-1.**

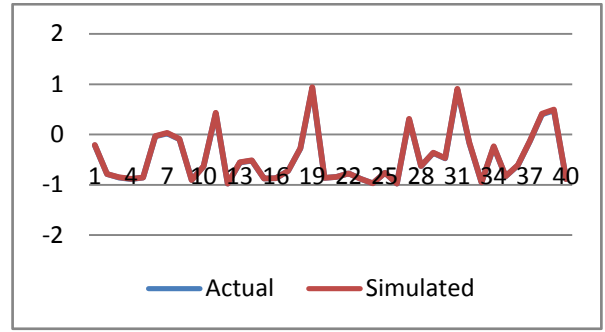

**Fig 4: Neural network output for Bohachevsky function-2.**

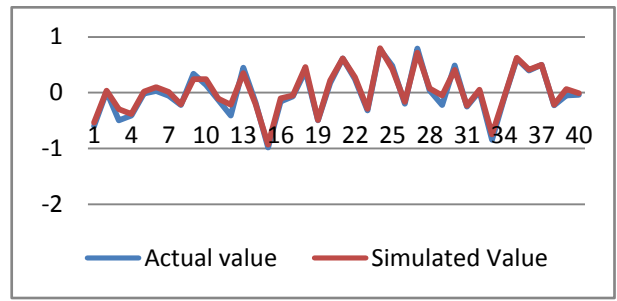

**Fig 5: Neural network output for Easom function.**

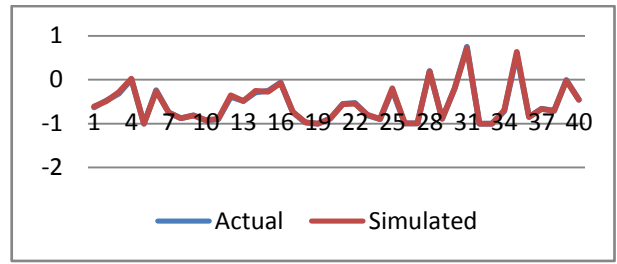

**Fig 6: Neural network output for Hump function.**

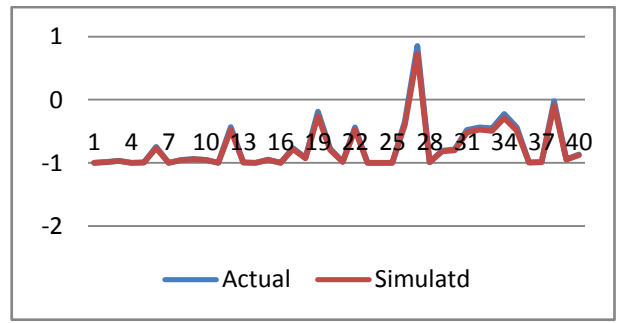

**Fig 7: Neural network output for Beale function.**

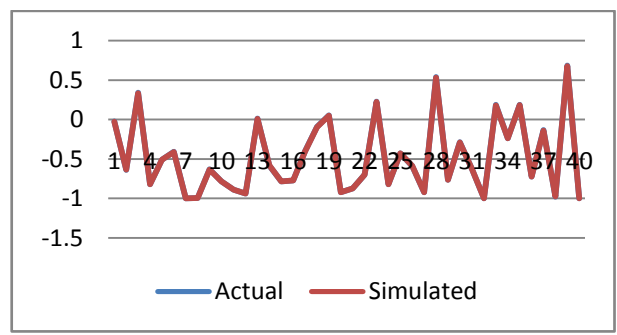

**Fig 8: Neural network output for Booth function.**

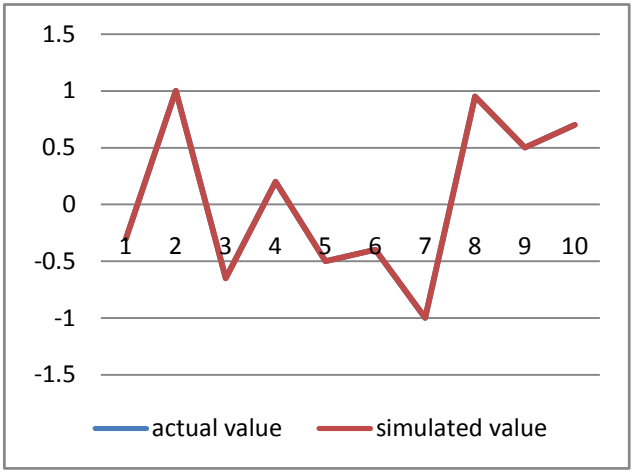

**Fig 9: Neural network output for Analog circuit function.**

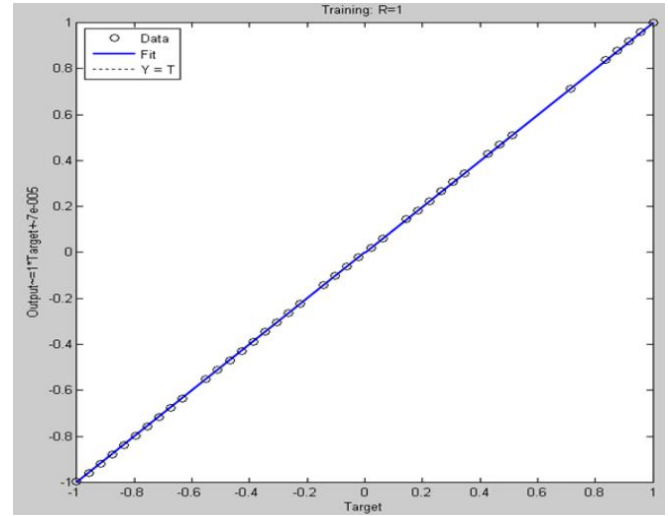

**Fig 12: Regression plot of ANN over analog circuit function.**

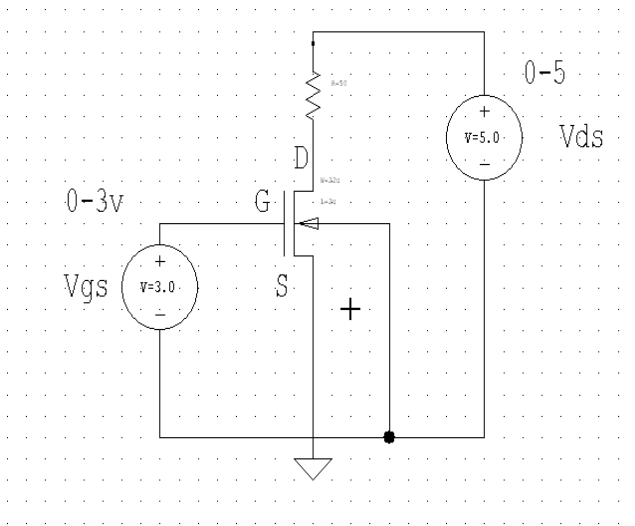

**Fig 10: Spice circuit diagram of an inverter**

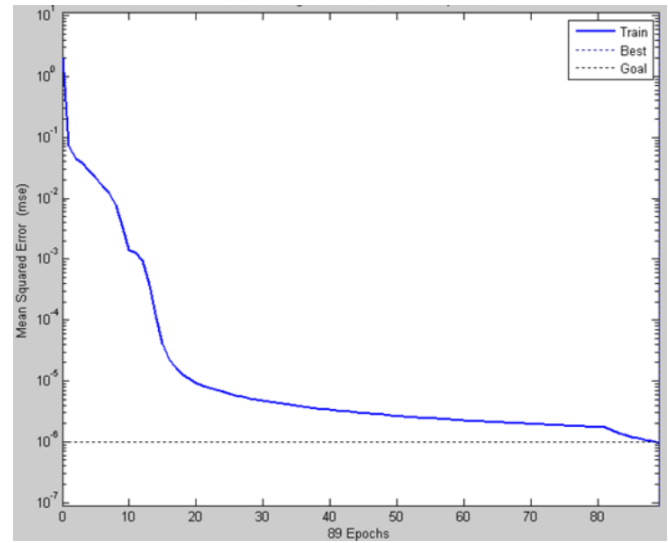

**Fig 13: Performance plot of ANN over analog circuit function.**

## 4. CONCLUSION

Increasing the number of hidden layers of neurons without effecting the speed factor can better the performance of the neural network. Reducing the mean square error can optimize the error. Synchronizing the network with much more learning rate can lead to more better performance. Handling the network with more data set, speed up and efficiency for much better performance.

Neural network modeling of analog circuit performance parameters proved to be an effective methodology for first and accurate performance estimation. Generating layout-aware models is one of the challenging tasks of analog macro modeling. Layout aware models are based on training data generated from simulations of circuit models extracted from layout. The use of better sampling methods to reduce a number of sample points can removed the drawback of accurately mapping the behavior of a circuit. Although a large number of simulations are required to capture the behavior of the performance characteristic of a circuit, the effort is justifiable when considering the reusability of the models. Performing on a particular experiment using SPICE would require much more timing evaluations where as the execution time of the same experiment using neural network provides a great deal of time savings in situations where a fixed topology must be reused and re-synthesized many times. The neural network models are also robust. Numerical stability in SPICE and other circuit simulators can prohibit the acquisition of performance parameters for some of the circuit configurations in the sample space. The ANN model can give estimate of values that the simulator failed to provide. However there are never any guarantees of absolute accuracy when approximating unknown functions. Use of a sufficiently large validation data set helps insure accuracy for most of the points in the input space. So neural networks have currently gained attention as a fast, accurate, reusable and flexible tool for analog circuit modeling simulation and design.

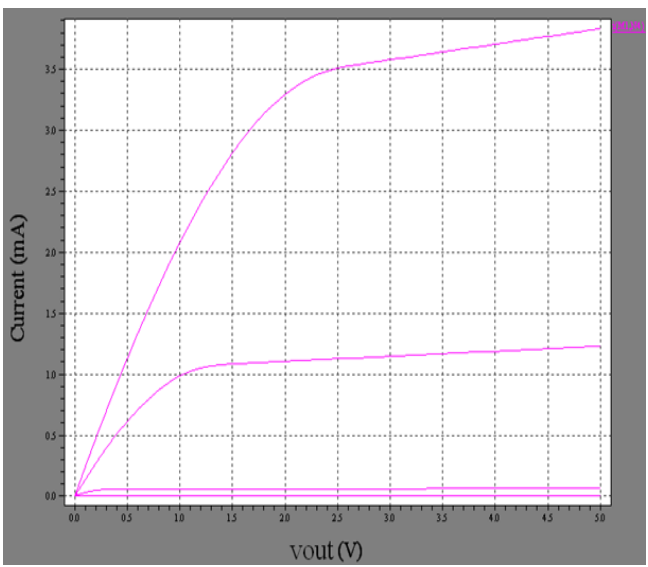

**Fig 11: Current voltage characteristics curve of a typical inverter circuit extracted from spice synthesis.**

## 5. ACKNOWLEDGMENT
The author thanks Dr. Rajib Chakraborty of the Department of Applied Optics and Photonics**,** University of Calcutta, Kolkata, West Bengal, India and Dr. Soumya Pandit of the department of Radio Physics and Electronics, University of Calcutta, Kolkata, West Bengal, India for their continuous support throughout the research work.

## 6. REFERENCES

[1] Neural Networks – A Comprehensive Foundation Simon Haykin.

[2] ANN for RF and Microwave Design-From theory to practice Q.J. Ziang and K.C. Gupta.

[3] B. Hassinbi, D. G. Stork, and G. J. Wolff, "Optimal brain surgeon and general network pruning," in Proc. IEEE Int. Joint Conf. Neural Netw., 1992, vol. 2, pp. 441–444.

[4] B. Widrow and R. Winter, "Neural nets for adaptive filtering and adap¬tive pattern recognition," Computer, vol. 21, no. 3, pp. 25–39, Mar. 1988

[5] K. Fukushima, S. Miyake, and T. Ito, "Neocognitron: A neural network model for a mechanism of visual pattern recognition," IEEE Trans. Syst., Man, Cybern. , vol. SMC-13, no. 5, pp. 826–834, 1983

[6] S. Grossberg, E. Mingolla, and D. Todorovic, "A neural network archi¬tecture for preattentive vision," IEEE Trans. Biomed. Eng., vol. 36, no. 1, pp. 65–84, Jan. 1989

[7] L. M. Reyneri, "Implementation issues of neuro-fuzzy hardware: Going towards HW/SW codesign," IEEE Trans. Neural Netw., vol. 14, no. 1, pp. 176–194, Jan. 2003.

[8] M. Cristea and A. Dinu, "A new neural network approach to induction motor speed control," in Proc. IEEE Power Electron. Specialist Conf., 2001, vol. 2, pp. 784–788.

[9] Y. J. Chen and D. Plessis, "Neural network implementation on a FPGA," in Proc. IEEE Africon Conf., 2002, vol. 1, pp. 337–342

[10] M. Marchesi, G. Orlandi, F. Piazza, and A. Uncini, "Fast neural net-works without multipliers," IEEE Trans. Neural Netw., vol. 4, no. 1, pp. 53–62, Jan. 1993

[11] B. Noory and V. Groza, "A reconfigurable approach to hardware im-plementation of neural networks," in Can. Conf. Electr. Comput. Eng., 2003, pp. 1861–1863

[12] J. Zhu, G. J. Milne, and B. K. Gunther, "Towards an FPGA based re-configurable computing environment for neural network implementa-tions," Inst. Elect. Eng. Proc. Artif. Neural Netw., vol. 2, no. 470, pp. 661–666, Sep. 1999.

[13] R. H. Turner and R. F. Woods, "Highly efficient limited range mul-tipliers for LUT-based FPGA architectures," IEEE Trans. Very Large Scale Integr. (VLSI) Syst., vol. 15, no. 10, pp. 1113–1117, Oct. 2004.

[14] K. M. Hornick, M. Stinchcombe, and H. white, "Multilayer feedfor-ward neural networks are universal approximators," Neural Netw., vol. 2, no. 5, pp. 141–154, 1985

[15] P. Vas, Sensorless Vector and Direct Torque Control. Oxford, U.K.: Oxford Univ. Press, 1998.

[16] V. Vapnik, Statistical Learning Theory. New York: Wiley, 1998.

[17] C. Bishop, Neural Networks for Pattern Recognition. Oxford, U.K.: Oxford Univ. Press, 1995

[18] F. L. P. Na, F. Bellas, R. Duro, and M. S. Simon, "Using adaptive artificial neural networks for reconstructing irregularly sampled laser doppler velocimetry signals," IEEE Trans. Instrum. Meas., vol. 55, no. 3, pp. 916–922, Jun. 2006

[19] C. Lin, "Training nu-support vector regression: Theory and algo-rithms," Neural Computation, vol. 14, pp. 1959–1977, 2002.

[20] M. Sorensen, "Functional consequences of model complexity in hy-brid neural-microelectronic systems," Ph.D. dissertation, Georgia Inst. Technol., Atlanta, 2005.

[21] M. L. Hines and N. T. Carnevale, "The NEURON simulation environ-ment," Neural Comp., vol. 9, no. 6, pp. 1179–1209, 1997.

[22] CMOS Analog Circuit Design – Phillip E. Allen, Douglas R. Holberg.

## 7. AUTHORS PROFILE
**Mriganka Chakraborty** was born on 11[th] October 1984 in Kolkata, West Bengal, India. He has received  the B.Tech degree in computer science & engineering and M.Tech degree in vlsi & microelectronics from West Bengal University of Technology, Kolkata, West Bengal, India in 2006 and 2009 respectively. He is currently an Asst. Professor with the department of Computer Science & Engineering in Seacom Engineering College, Howrah, West Bengal, India. His research interests includes ANN, VLSI circuit design, Network On Chip, Physical VLSI Design techniques.